\documentclass[11pt,a4paper]{article}
\usepackage{authblk}
\usepackage[hyperref]{acl2020}
\usepackage{times}
\usepackage{latexsym}

\usepackage{amsmath}
\usepackage{amssymb}

\usepackage{url}
\usepackage{booktabs} % For formal tables
\usepackage{graphicx}
\usepackage{epstopdf}
\usepackage{subfigure}
\usepackage{verbatim}
\usepackage{amssymb}
\usepackage{amsmath}
\usepackage{bm}
\usepackage{array}
\usepackage{multirow}
\usepackage{subfigure}
\usepackage[most]{tcolorbox}

\usepackage[font={small}]{caption}

\usepackage{microtype}

\aclfinalcopy % Uncomment this line for the final submission
\title{Few-Shot NLG with Pre-Trained Language Model}

\author[1]{\textbf{Zhiyu Chen}}
\author[2]{\textbf{Harini Eavani}}
\author[1]{\textbf{Wenhu Chen}}
\author[2]{\textbf{Yinyin Liu}}
\author[1]{\textbf{William Yang Wang}}
\affil[1]{University of California, Santa Barbara}
\affil[2]{Intel AI \authorcr \{zhiyuchen, wenhuchen, william\}@cs.ucsb.edu, \{harini.eavani, yinyin.liu\}@intel.com}

\date{}

\begin{document}
\maketitle
\begin{abstract}
Neural-based end-to-end approaches to natural language generation (NLG) from structured data or knowledge are data-hungry, making their adoption for real-world applications difficult with limited data. In this work, we propose the new task of \textit{few-shot natural language generation}. Motivated by how humans tend to summarize tabular data, we propose a simple yet effective approach and show that it not only demonstrates strong performance but also provides good generalization across domains. The design of the model architecture is based on two aspects: content selection from input data and language modeling to compose coherent sentences, which can be acquired from prior knowledge. With just 200 training examples, across multiple domains, we show that our approach achieves very reasonable performances and outperforms the strongest baseline by an average of over 8.0 BLEU points improvement. Our code and data can be found at \url{https://github.com/czyssrs/Few-Shot-NLG}
\end{abstract}
\section{Introduction}
\begin{figure*}[htbp]
\begin{minipage}[t]{0.28\textwidth}
\centering
\vspace{10pt}
\resizebox{0.8\textwidth}{!}{%
\begin{tabular}{ll}
% \toprule
Input Table \\
\toprule
Attribute (R) & Value (V) \\
\midrule
Name & Walter Extra \\
Nationality & German \\
Occupation & Aircraft designer \\
 & and manufacturer \\
... & ...\\
\bottomrule
\end{tabular}
}
\end{minipage}
\begin{minipage}[t]{0.7\textwidth}
\centering
\vspace{-10pt}
\includegraphics[width=4.0in]{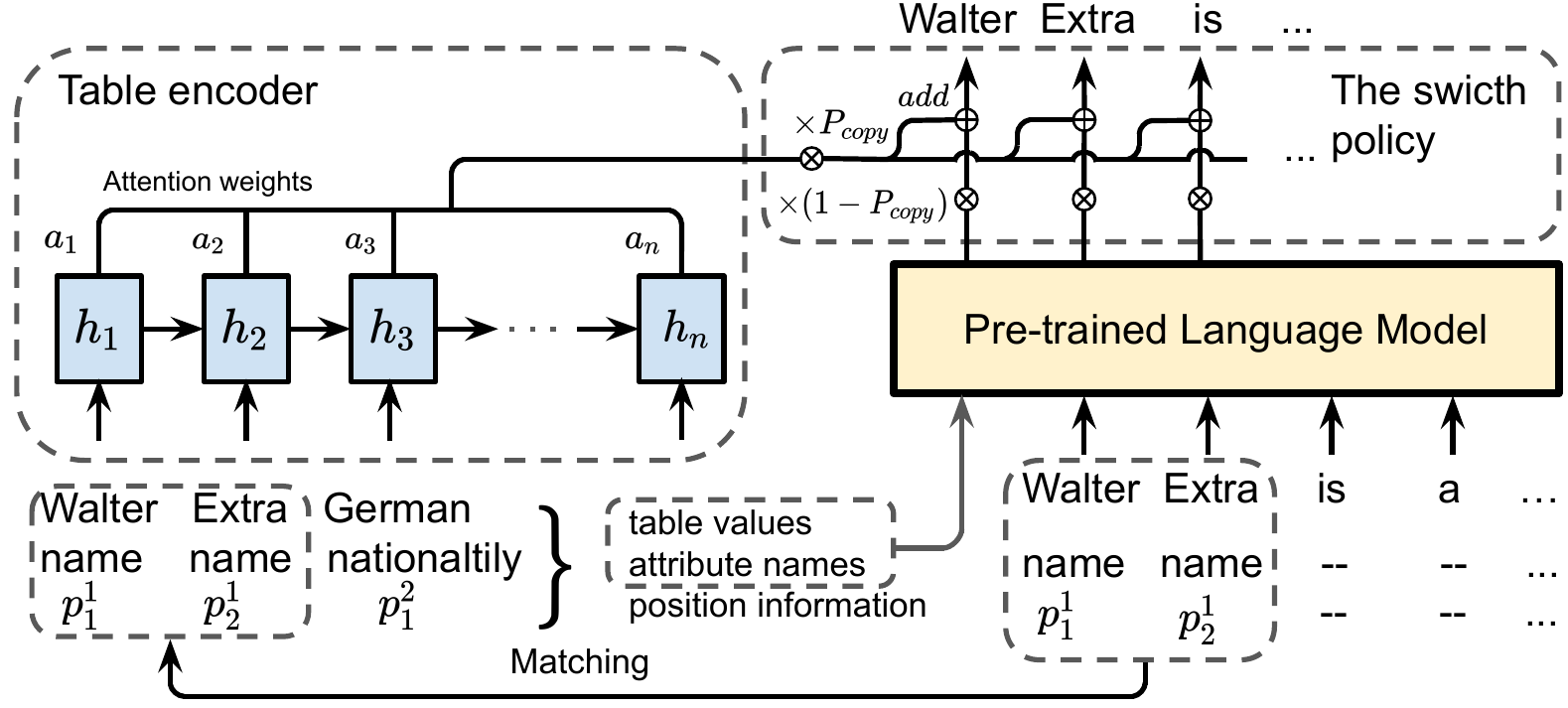}
\end{minipage}
\caption{Overview of our approach: Under the base framework with switch policy, the pre-trained language model serves as the generator. We follow the same encoder as in~\cite{DBLP:conf/aaai/LiuWSCS18}. The architecture is simple in terms of both implementation and parameter space that needs to be learned from scratch, which should not be large given the few-shot learning setting.} 
\label{fig:model}
\end{figure*}

Natural language generation (NLG) from structured data or knowledge ~\cite{gatt2018survey} is an important research problem for various NLP applications. Some examples are task-oriented dialog, question answering~\cite{DBLP:conf/acl/HeBEL17, DBLP:conf/aaai/GhazvininejadBC18,DBLP:conf/emnlp/SuSSSGYY16, DBLP:conf/aaai/SahaPKSC18,DBLP:conf/ijcai/YinJLSLL16} and interdisciplinary applications such as medicine~\cite{hasan2019clinical,cawsey1997natural} and health-care~\cite{hasan2019clinical,dimarco2007development}. There is great potential to use automatic NLG systems in a wide range of real-life applications.
%, with applications in task-oriented dialog, question answering~\cite{DBLP:conf/acl/HeBEL17, DBLP:conf/aaai/GhazvininejadBC18,DBLP:conf/emnlp/SuSSSGYY16} and interdisciplinary applications such as medicine~\cite{hasan2019clinical} and health-care~\cite{dimarco2007development}.
%Prior work in this area includes pipeline-based NLG systems~\cite{becker2002practical,deemter2005real,gatt2018survey}, which heavily rely on template and slot-filling rules to produce very accurate but limited text output. Such a pipeline-based paradigm~\cite{DBLP:journals/nle/ReiterD97} divides the generation process into parts, optimizing separately for each objective such as content selection followed by macro/micro-planning and surface realization. 
Recently, deep neural network based NLG systems have been developed, such as those seen in the E2E challenge~\cite{DBLP:conf/sigdial/NovikovaDR17}, \textsc{WeatherGov}~\cite{DBLP:conf/acl/LiangJK09}, as well as more complex ones such as \textsc{WikiBio}~\cite{DBLP:conf/aaai/LiuWSCS18} and \textsc{RotoWire}~\cite{DBLP:conf/emnlp/WisemanSR17}. Compared to traditional slot-filling pipeline approaches, such neural-based systems greatly reduce feature engineering efforts and improve text diversity as well as fluency.  

Although they achieve good performance on benchmarks such as E2E challenge~\cite{DBLP:conf/sigdial/NovikovaDR17} and \textsc{WikiBio}~\cite{DBLP:conf/emnlp/LebretGA16}, 
% their methods such as~\cite{DBLP:conf/aaai/LiuWSCS18} is based on training strong domain-specific language model, 
their performance depends on large training datasets, e.g., 500k table-text training pairs for \textsc{WikiBio}~\cite{DBLP:conf/emnlp/LebretGA16} in a single domain. Such data-hungry nature makes neural-based NLG systems difficult to be widely adopted in real-world applications as they have significant manual data curation overhead. This leads us to formulate an interesting research question:
\begin{quote}
    \emph{1. Can we significantly reduce human annotation effort to achieve reasonable performance using neural NLG models?}\\
    \emph{2. Can we make the best of generative pre-training, as prior knowledge, to generate text from structured data?}
\end{quote}
Motivated by this, we propose the new task of \emph{few-shot natural language generation}: given only a handful of labeled instances (e.g., 50 - 200 training instances), the system is required to produce satisfactory text outputs (e.g., BLEU $\geq$ 20). To the best of our knowledge, such a problem in NLG community still remains under-explored. Herein, we propose a simple yet very effective approach that can generalize across different domains.

In general, to describe information in a table, we need two skills to compose coherent and faithful sentences. One skill is to select and copy factual content from the table - this can be learned quickly by reading a handful of tables. The other is to compose grammatically correct sentences that bring those facts together - this skill is not restricted to any domain. One can think of a latent ``switch'' that helps us alternate between these two skills to produce factually correct and coherent sentences.
To do this, we use the pre-trained language model~\cite{chelba2013one,radford2019language} as the innate language skill, which provides strong prior knowledge on how to compose fluent and coherent sentences. The ability to switch and select/copy from tables can be learned successfully using only a few training instances, freeing the neural NLG model from data-intensive training. 
Previous best performing methods based on large training data, such as~\cite{DBLP:conf/aaai/LiuWSCS18}, which does not apply such switch mechanism but trains a strong domain-specific language model, perform very poorly under few-shot setting.

Since we are operating under a highly data-restricted few-shot regime, we strive for simplicity of model architecture. This simplicity also implies better generalizability and reproducibility for real-world applications. 
% We follow the method laid out in \textsc{WikiBio}~\cite{DBLP:conf/emnlp/LebretGA16} to
We crawl multi-domain table-to-text data from Wikipedia as our training/test instances.
With just 200 training instances, our method can achieve very reasonable performance.

In a nutshell, our contributions are summarized as the following:
\begin{itemize}
    \item We propose the new research problem of few-shot NLG, which has great potential to benefit a wide range of real-world applications. 
    \item To study different algorithms for our proposed problem, we create a multi-domain table-to-text dataset.
    \item Our proposed algorithm can make use of the external resources as prior knowledge to significantly decrease human annotation effort and improve the baseline performance by an average of over 8.0 BLEU on various domains.
\end{itemize}

\section{Related Work}
\subsection{NLG from Structured Data}
As it is a core objective in many NLP applications, natural language generation from structured data/knowledge (NLG) has been studied for many years. Early traditional NLG systems follow the pipeline paradigm that explicitly divides generation into content selection, macro/micro planning and surface realization~\cite{DBLP:journals/nle/ReiterD97}. Such a pipeline paradigm largely relies on templates and hand-engineered features. Many works have been proposed to tackle the individual modules, such as~\cite{DBLP:conf/acl/LiangJK09,DBLP:conf/naacl/WalkerRR01,DBLP:conf/emnlp/LuNL09}. Later works~\cite{DBLP:conf/naacl/KonstasL12,DBLP:journals/jair/KonstasL13} investigated modeling context selection and surface realization in an unified framework. 

Most recently, with the success of deep neural networks, data-driven, neural based approaches have been used, including the end-to-end methods that jointly model context selection and surface realization~\cite{DBLP:conf/aaai/LiuWSCS18,DBLP:conf/emnlp/WisemanSR18,DBLP:journals/corr/abs-1809-00582}. Such data-driven approaches achieve good performance on several benchmarks like E2E challenge~\cite{DBLP:conf/sigdial/NovikovaDR17}, WebNLG challenge~\cite{DBLP:conf/inlg/GardentSNP17} and \textsc{WikiBio}~\cite{DBLP:conf/emnlp/LebretGA16}. However, they rely on massive amount of training data. ElSahar et al.~\shortcite{DBLP:conf/naacl/ElSaharGL18} propose zero-shot learning for question generation from knowledge graphs, but their work applies on the transfer learning setting for unseen knowledge base types, based on seen ones and their textual contexts, which still requires large in-domain training dataset. This is different from our few-shot learning setting. Ma et al.~\shortcite{DBLP:conf/acl/MaYLLZS19} propose low-resource table-to-text generation with 1,000 paired examples and large-scale target-side examples. In contrast, in our setting, only tens to hundreds of paired training examples are required, meanwhile without the need for any target examples. This is especially important for real-world use cases where such large target-side gold references are mostly hard to obtain. Therefore, our task is more challenging and closer to real-world settings.

\subsection{Large Scale Pre-Trained Models}
Many of the current best-performing methods for various NLP tasks adopt a combination of pre-training followed by supervised fine-tuning, using task-specific data. Different levels of pre-training include word embeddings~\cite{DBLP:conf/nips/MikolovSCCD13,DBLP:conf/emnlp/PenningtonSM14,DBLP:conf/naacl/PetersNIGCLZ18}, sentence embeddings~\cite{DBLP:conf/icml/LeM14,DBLP:conf/nips/KirosZSZUTF15}, and most recently, language modeling based pre-training like BERT~\cite{devlin2018bert} and GPT-2~\cite{radford2019language}. Such models are pre-trained on large-scale open-domain corpora, and provide down-streaming tasks with rich prior knowledge while boosting their performance. In this paper, we adopt the idea of employing a pre-trained language model to endow in-domain NLG models with language modeling ability, which cannot be well learned from few shot training instances.
\section{Method}
\begin{table*}[h]
\begin{center}
\resizebox{1.0\textwidth}{!}{%
\begin{tabular}{lccccccccccccccccccccccc}
\toprule
Domain &  &
\multicolumn{5}{c}{Humans} &  &  &  &
\multicolumn{5}{c}{Books} &  &  &  &
\multicolumn{5}{c}{Songs}\\
\cline{3-7}
\cline{11-15}
\cline{19-23}
\# of training instances &  & - & 50 & 100 & 200 & 500 &  &  &  & - & 50 & 100 & 200 & 500 &  &  &  & - & 50 & 100 & 200 & 500\\
\midrule
Template &  & 16.3 & - & - & - & - &  &  &  & 25.6 & - & - & - & - &  &  &  & 30.1 & - & - & - & - \\
\midrule
Base-original &  & - & 2.2 & 3.7 & 4.9 & 5.1 &  &  &  & - & 5.8 & 6.1 & 7.4 & 6.7 & &  &  & - & 9.2 & 10.7 & 11.1 & 11.3 \\
Base &  & - & 2.9 & 5.1 & 6.1 & 8.3 &  &  &  & - & 7.3 & 6.8 & 7.8 & 8.8 &  &  &  & - & 10.4 & 12.0 & 11.6 & 13.1\\
\midrule
Base + switch &  & - & 15.6 & 17.8 & 21.3 & 26.2 &  &  &  & - & 24.7 & 26.9 & 30.5 & 33.2 &  &  &  & - & 29.7 & 30.6 & 32.5 & 34.9 \\
Base + switch + LM-scratch &  & - & 6.6 & 11.5 & 15.3 & 18.6 &  &  &  & - & 7.1 & 9.2 & 14.9 & 21.8 &  &  &  & - & 11.6 & 16.2 & 20.6 & 23.7\\
Base + switch + LM (Ours) &  & - & \textbf{25.7} & \textbf{29.5} & \textbf{36.1} & \textbf{41.7} &  &  &  & - & \textbf{34.3} & \textbf{36.2} & \textbf{37.9} & \textbf{40.3} &  &  &  & - & \textbf{36.1} & \textbf{37.2} & \textbf{39.4} & \textbf{42.2}\\
\bottomrule
\end{tabular}
}
\end{center}
\caption{BLEU-4 results on three domains. Base-original: the original method in~\cite{DBLP:conf/aaai/LiuWSCS18}; Base: applies pre-trained word embedding; Base+switch: adds the switch policy; Base+switch+LM-scratch: makes the same architecture as our method, but trains the model from scratch without pre-trained weights for the generator. Template: manually crafted templates}
\label{table:res_bleu}
\end{table*}
\label{sec:method}
\subsection{Problem Formulation}
\label{sec:background}
We are provided with semi-structured data: a table of attribute-value pairs $\{R_i:V_i\}_{i=1}^{n}$. Both $R_i$ and $V_i$ can be either a string/number, a phrase or a sentence. Each value is represented as a sequence of words $V_i = \{v_j\}_{j=1}^{m}$.
For each word $v_j$, we have its corresponding attribute name $R_i$ and position information of the word in the value sequence.
The target is to generate a natural language description based on the semi-structured data, provided with only a handful of training instances.
\subsection{Base Framework with Switch Policy}
\label{sec:switch}
We start with the field-gated dual attention model proposed in~\cite{DBLP:conf/aaai/LiuWSCS18}, which achieves state-of-the-art performance (BLEU) on \textsc{WikiBio} dataset. Their method uses an LSTM decoder with dual attention weights.
We first apply a switch policy that decouples the framework into table content selection/copying and language model based generation. 
Inspired by the pointer generator~\cite{DBLP:conf/acl/SeeLM17}, at each time step, we maintain a soft switch $p_{copy}$ to choose between generating from softmax over vocabulary or copying from input table values with the attention weights as the probability distribution. 
\begin{equation}\nonumber
\begin{aligned}
    &p_{copy} = \text{sigmoid}(W_cc_t + W_ss_t + W_xx_t + b)
\end{aligned}
\end{equation}
Where $c_t = \sum_{i}a_t^ih_i$, $\{h_i\}$ is the encoder hidden states, $x_t, s_t, a_t$ is the decoder input, state and attention weights respectively at time step $t$. $W_c, W_s, W_x$ and $b$ are trainable parameters. 

% difference between pointer generator
The pointer generator learns to alternate between copying and generating based on large training data and shows its advantage of copying out-of-vocabulary words from input. In our task, the training data is very limited, and many of the table values are not OOV. We need to explicitly ``teach'' the model where to copy and where to generate. Therefore, to provide the model accurate guidance of the behavior of the switch, we match the target text with input table values to get the positions of where to copy. At these positions, we maximize the copy probability $p_{copy}$ via an additional loss term. Our loss function:
\begin{equation}\nonumber
\begin{aligned}
    &L = L_c + \lambda\sum_{\begin{subarray}{c}w_{j} \in m \\ m \in \{V_i\}\end{subarray}} ( 1 - p^j_{copy} )
\end{aligned}
\end{equation}
Where $L_c$ is the original loss between model outputs and target texts. $w_j$ is the target token at position $j$, $\{V_i\}$ is the input table value list defined in Section \ref{sec:background}, and $m$ means a matched phrase. $\lambda$ is hyperparameter as the weight for this copy loss term.
We also concatenate the decoder input with its matched attribute name and position information in the input table as $x_t$ to calculate $p_{copy}$ . 
%See figure~\ref{fig:model-switch} for an overview of the base framework with the switch policy. 
%\begin{figure}[htbp]
%\begin{center}
%	\includegraphics[width=.48\textwidth]{figs/model-switch.pdf}
%    \caption{The base framework with switch policy: We follow the same encoder architecture as in~\cite{DBLP:conf/aaai/LiuWSCS18}. During training, we match the target text with input table values to get the attribute name and position information of the tokens that should be copied. During testing, if the last token is copied from input table, then we get its attribute name and position information from corresponding table slots.}
%    \label{fig:model-switch}
%\end{center}
%\end{figure}
\subsection{Pre-Trained LM as Generator}
\label{sec:lm}
% We replace the LSTM decoder in the base framework we investigate above, with a pre-trained language model. 

%Such language models are typically trained using a large, open-domain corpus. In the absence of any specific constraint or in-domain orientation, the pool of candidate words at each generation step will be quite large. Given an NLG task for a specific domain, we need to instill the in-domain language modeling paradigm, to narrow down the candidate pool. We do this using fine-tuning, as described below.
%This is motivated by human speech - humans have many speech ``modes''; for example, the talking mode when writing a research paper differs from the talking mode when chatting with friends. While both come from the same general innate talking ability, a domain-specific mode (such as the use of technical terms in a research talk, or the use of informal language in a conversation with friends) is selected based on the situation encountered. In our task, for example, when generating biography for human beings, a sample language paradigm could be "$<$name$>$, born on $<$birth\_date$>$, is a $<$nationality$>$ $<$occupation$>$ ...". When generating description for a book, a sample language paradigm could be "$<$book\_title$>$ is a novel written by $<$author$>$, published in ...". Both paradigms are embedded in the pre-trained language model.
% We tailor a pre-trained language model to a specific domain using fine-tuning with a few in-domain training instances. 
We use a pre-trained language model as the generator, serving as the ``innate language skill''. 
% We distill the language modeling paradigm of a specific domain through fine-tuning.
Due to the vocabulary limitation of few training instances, we leave the pre-trained word embedding fixed while fine-tuning other parameters of the pre-trained language model, so that it can generalize with tokens unseen during training. 

Figure \ref{fig:model} shows our model architecture. We use the pre-trained language model GPT-2\footnote{https://github.com/openai/gpt-2} proposed in ~\cite{radford2019language}, which is a 12-layer transformer. The final hidden state of the transformer is used to calculate attention weights and the copy switch $p_{copy}$.
% Note that as the transformer architecture does not have an initial state, 
We first feed the embedded attribute-value list serving as the context for generation. In this architecture, the generator is fine-tuned from pre-trained parameters while the encoder and attention part is learned from scratch, the initial geometry of the two sides are different. Therefore we need to apply larger weight to the copy loss $p_{copy}$, to give the model a stronger signal to ``teach" it to copy facts from the input table. 
\section{Experiment}
\subsection{Datasets and Experiment Setup}
The original \textsc{WikiBio} dataset~\cite{DBLP:conf/emnlp/LebretGA16} contains 700k English Wikipedia articles of well-known humans, with the Wiki infobox serving as input structured data and the first sentence of the article serving as target text. To demonstrate generalizability, we collect datasets from two new domains: \textit{Books} and \textit{Songs} by crawling Wikipedia pages. After filtering and cleanup, we end up with 23,651 instances for \textit{Books} domain and 39,450 instances for \textit{Songs} domain\footnote{Note that the target text sometimes contains information not in the infobox. This is out of the scope of the few-shot generation in this work. Therefore we further filter the datasets and remove the ones with rare words out of infobox. Check~\cite{DBLP:conf/acl/DhingraFPCDC19} for a related study of this issue on the WikiBio dataset}.
Together with the \textit{Humans} domain of the original \textsc{WikiBio} dataset, for all three domains we conduct experiments by varying the training dataset size to 50, 100, 200 and 500. The rest of data is used for validation (1,000) and testing. The weight $\lambda$ of the copy loss term is set to 0.7. Other parameter settings can be found in Appendix A. To deal with vocabulary limitation of few-shot training, for all models we adopt the Byte Pair Encoding (BPE)~\cite{DBLP:conf/acl/SennrichHB16a} and subword vocabulary in~\cite{radford2019language}. 

We compare the proposed method with other approaches investigated in Section \ref{sec:method}, serving as the baselines - \textbf{Base-original: } the original model in~\cite{DBLP:conf/aaai/LiuWSCS18}; \textbf{Base: } uses the same architecture, but in addition applies the pre-trained word embedding and fix it during training; \textbf{Base + switch: } adds the switch policy; \textbf{Base + switch + LM-scratch: } makes the architecture same as our method, except training the model from scratch instead of using pre-trained weights for generator. \textbf{Template: } template-based non-neural approach, manually crafted for each domain. 
\subsection{Results and Analysis}
Following previous work~\cite{DBLP:conf/aaai/LiuWSCS18}, we first conduct automatic evaluations using BLEU-4, shown in Table~\ref{table:res_bleu}. The ROUGE-4 (F-measure) results follow the same trend with BLEU-4 results, which we show in Appendix B. 

As we can see, the original model \textbf{Base-original}~\cite{DBLP:conf/aaai/LiuWSCS18}, which obtains the state-of-the-art result on \textsc{WikiBio} full set, performs very poorly under few-shot setting. It generates all tokens from softmax over vocabulary, which results in severe overfitting with limited training data, and the results are far behind the template-based baseline. With the switch policy, \textbf{Base+switch} first brings an improvement of an average of over 10.0 BLEU points. This indicates that the content selection ability is easier to be learned with a handful of training instances. However, it forms very limited, not fluent sentences. With the augmentation of the pre-trained language model, our model \textbf{Base+switch+LM} brings one more significant improvement of an average over 8.0 BLEU points. 
We provide sample outputs of these methods using 200 training instances in Table~\ref{table:case_study}. 
\begin{table}[htbp]
\begin{center}
\resizebox{.48\textwidth}{!}{%
\begin{tabular}{llll}
\toprule
Attribute & Value & Attribute & Value\\
\midrule
name & \colorbox{blue!15}{andri ibo} & fullname & andri ibo\\
birth date & \colorbox{yellow!30}{3 april 1990} & birth place & sentani , jayapura , \colorbox{orange!20}{indonesia}\\
height & 173 cm & currentclub & \colorbox{green!15}{persipura jayapura}\\
position & \colorbox{cyan!15}{defender} & ... \\
% name & \colorbox{blue!15}{andri ibo} \\
% fullname & andri ibo \\
% birth date & \colorbox{yellow!30}{3 april 1990} \\
% birth place & sentani , jayapura , \colorbox{orange!20}{indonesia} \\
% height & 173 cm \\
% currentclub & \colorbox{green!15}{persipura jayapura} \\
% position & \colorbox{cyan!15}{defender} \\
% youthyears & 2008 -- 2009 \\
% youthclubs & persidafon u-21 \\
% years & 2008 -- 2009  2013 --  \\
% clubs & persidafon u-21 persipura jayapura \\
% caps & 56 8 \\  
% goals & 2 0 \\
% nationalyears & 2013 \\
% nationalteam & indonesia  u-23 \\
% nationalcaps & 8 \\   
% nationalgoals & 2 \\
% pcupdate & 11 november 2014 \\
% article title & andri ibo \\
% \midrule
% \multicolumn{4}{c}{Generated Texts} \\
\midrule
\multicolumn{4}{l}{
\begin{minipage}[t]{1.5\columnwidth}
\textbf{Gold Reference}: \colorbox{blue!15}{andri ibo} ( born \colorbox{yellow!30}{april 3 , 1990} ) is an \colorbox{orange!20}{indonesian} footballer who currently plays for \colorbox{green!15}{persipura jayapura} in the indonesia super league .
\end{minipage}}\\
\midrule
\multicolumn{4}{c}{Generated texts of different methods} \\
\midrule
\multicolumn{4}{l}{
\begin{minipage}[t]{1.5\columnwidth}
\textbf{Base}: vasco emanuel freitas ( born december 20 , 1992 in kong kong ) is a hong kussian football player and currently plays for hong kong first division league side tsw pegasus . 
\end{minipage}}\\
\midrule
\multicolumn{4}{l}{
\begin{minipage}[t]{1.5\columnwidth}
\textbf{Base+switch}: \colorbox{blue!15}{andri ibo} andri ibo ( \colorbox{yellow!30}{3 april 1990} ) is a international cricketer . 
\end{minipage}}\\
\midrule
\multicolumn{4}{l}{
\begin{minipage}[t]{1.5\columnwidth}
\textbf{Base+switch+LM (Ours)}: \colorbox{blue!15}{andri ibo} ( born \colorbox{yellow!30}{3 april 1990} ) is an \colorbox{orange!20}{indonesian} football \colorbox{cyan!15}{defender} , who currently plays for \colorbox{green!15}{persipura jayapura} . 
\end{minipage}}\\
\bottomrule
\end{tabular}
}
\end{center}
\caption{A sample input table and generated summaries from the test set of \textit{Humans} domain, using 200 training instances}
\label{table:case_study}
\end{table}
% The difference between \textbf{FG+switch} and \textbf{FG+switch+LM-scratch} is the architecture of the generator, with a 1-layer LSTM for the former one and 12-layer transformer for the latter. The generator of both methods is learned from scratch, while the transformer structure is more complex in terms of parameter space. Therefore its performance is inferior due to limited training data. 
% To study the robustness of the approach, we further train with 800 and 1000 instances, and plot the overall performance curve in figure~\ref{fig:curve}. Compare to the strongest baseline, FG+switch, the performance of our approach tends to increase faster at first (the region less than 200 instances), then gradually goes steady with the increase of training instances.  
% \subsection*{Ablation Study: Effect of the Switch Policy}

Table~\ref{table:copy_loss} shows the effect of the copy switch loss $p_{copy}$ introduced in Section \ref{sec:switch}, giving the model a stronger signal to learn to copy from input table. 
% The loss term brings an average improvement of over 4.0 BLEU points.
% Compared to the baseline framework FG+switch, this is more important for our model with an unbalanced initial geometry for the encoder side and generator side, since the latter is fined-tuned from pre-trained weights. Without a stronger signal to "teach" the model to learn to copy from inputs, the generator tends to be optimized into local minima and produce incorrect facts. 
% Table \ref{table:copy_loss} shows our results for \textit{Humans} domain, comparing with not using the copy loss term. 
\begin{table}[htbp]
\small
\begin{center}
\resizebox{.36\textwidth}{!}{%
\begin{tabular}{lcccc}
\toprule
\# of training instances & 50 & 100 & 200 & 500\\
\midrule
Base + switch + LM & \textbf{25.7} & \textbf{29.5} & \textbf{36.1} & \textbf{41.7}\\
 - w/o copy loss $p_{copy}$ & 21.4 & 25.5 & 31.3 & 38.0\\
\bottomrule
\end{tabular}
}
\caption{Ablation study: Effect of the copy loss term on \textit{Humans} domain, measured by BLEU-4. The loss term brings an average improvement of over 4.0 BLEU points.}
\label{table:copy_loss}
\end{center}
\end{table}

Ma et al.~\shortcite{DBLP:conf/acl/MaYLLZS19} propose the Pivot model, for low-resource NLG with 1,000 paired examples and large-scale target-side examples. We compare our method with the Pivot model in table~\ref{table:pivot}. Note that here we train and evaluate the models on the original WikiBio dataset used in their work, in order to maintain the size of the target side examples for their settings.
\begin{table}[htbp]
\small
\begin{center}
\resizebox{.48\textwidth}{!}{%
\begin{tabular}{lccccc}
\toprule
\# of paired training instances & 50 & 100 & 200 & 500 & 1000\\
\midrule
Pivot & 7.0 & 10.2 & 16.8 & 20.3 & 27.3\\
Ours & 17.2 & 23.8 & 25.4 & 28.6 & 31.2\\
\bottomrule
\end{tabular}
}
\caption{Comparison with the Pivot model~\cite{DBLP:conf/acl/MaYLLZS19}. Compared to their method using additional large-scale target side examples, our method requires no additional target side data, while achieving better performance.}
\label{table:pivot}
\end{center}
\end{table}

\subsection*{Human Evaluation}
We also conduct human evaluation studies using Amazon Mechanical Turk, based on two aspects: \textit{Factual correctness} and \textit{Language naturalness}. 
We evaluate 500 samples. Each evaluation unit is assigned to 3 workers to eliminate human variance. 
The first study attempts to evaluate how well the generated text correctly conveys information in the table, by counting the number of facts in the text supported by the table, and contradicting with or missing from the table.
The 2nd and 3rd columns of Table~\ref{table:manual} show the average number of supporting and contradicting facts for our method, comparing to the strongest baseline and the gold reference. 
The second study evaluates whether the generated text is grammatically correct and fluent, regardless of factual correctness. We conduct pairwise comparison among all methods, and calculate the average times each method is chosen to be better than another, shown in the 4th column of Table ~\ref{table:manual}.
Our method brings a significant improvement over the strongest baseline ($p < 0.01$ in Tukey's HSD test for all measures). The copy loss term further alleviates producing incorrect facts. The language naturalness result of our method without the copy loss is slightly better, because this evaluation does not consider factual correctness; thus the generated texts with more wrong facts can still get high score.
See Appendix C for more details of our evaluation procedure.
% texts generated without the copy loss term can still get high score even though have more wrong facts.
% Each time a generated text is chosen as the better one, we add score of 1.0 to the corresponding method. If two texts are tied, we add 0.5 for each method. We show the average score for each method in the forth column of table~\ref{table:manual}, indicating its superiority in pairwise comparisons with all other methods.
\begin{table}[htbp]
\begin{center}
\resizebox{.41\textwidth}{!}{%
\begin{tabular}{lcccc}
\toprule
 & \# Supp. & \# Cont. &  & Lan. Score\\
\cline{2-3}
\cline{5-5}
Gold Reference & 4.25 & 0.84 &  & 1.85\\
\midrule
Base + switch & 2.57 & 2.17 &  & 0.93\\
\midrule
Base + switch + LM (ours) & \textbf{3.64} & \textbf{1.12} &  & 1.59\\
 - w/o copy loss $p_{copy}$ & 3.54 & 1.30 &  & \textbf{1.63}\\
\bottomrule
\end{tabular}
}
% ??? explain the metrics. larger or smaller the better?
\caption{Human evaluation results: Average number of supporting facts (column 2, the larger the better), contradicting facts (column 3, the smaller the better), and language naturalness score (column 4, the larger the better).}
\label{table:manual}
\end{center}
\end{table}
\section{Conclusion}
In this paper, we propose the new research problem of few-shot natural language generation. Our approach is simple, easy to implement, while achieving strong performance on various domains.  Our basic idea of acquiring language modeling prior can be potentially extended to a broader scope of generation tasks, based on various input structured data, such as knowledge graphs, SQL queries, etc. The deduction of manual data curation efforts for such tasks is of great potential and importance for many real-world applications.

\section*{Acknowledgment}
We thank the anonymous reviewers for their thoughtful comments. We thank Shuming Ma for releasing the processed data and code for the Pivot model.
This research was supported by the Intel AI Faculty Research Grant. The authors are solely responsible for the contents of the paper and the opinions expressed in this publication do not reflect those of the funding agencies.

\clearpage

\bibliography{acl2020}

\begin{thebibliography}{36}
\expandafter\ifx\csname natexlab\endcsname\relax\def\natexlab#1{#1}\fi

\bibitem[{Cawsey et~al.(1997)Cawsey, Webber, and Jones}]{cawsey1997natural}
Alison~J Cawsey, Bonnie~L Webber, and Ray~B Jones. 1997.
\newblock Natural language generation in health care.

\bibitem[{Chelba et~al.(2013)Chelba, Mikolov, Schuster, Ge, Brants, Koehn, and
  Robinson}]{chelba2013one}
Ciprian Chelba, Tomas Mikolov, Mike Schuster, Qi~Ge, Thorsten Brants, Phillipp
  Koehn, and Tony Robinson. 2013.
\newblock One billion word benchmark for measuring progress in statistical
  language modeling.
\newblock \emph{arXiv preprint arXiv:1312.3005}.

\bibitem[{Devlin et~al.(2018)Devlin, Chang, Lee, and
  Toutanova}]{devlin2018bert}
Jacob Devlin, Ming-Wei Chang, Kenton Lee, and Kristina Toutanova. 2018.
\newblock Bert: Pre-training of deep bidirectional transformers for language
  understanding.
\newblock \emph{arXiv preprint arXiv:1810.04805}.

\bibitem[{Dhingra et~al.(2019)Dhingra, Faruqui, Parikh, Chang, Das, and
  Cohen}]{DBLP:conf/acl/DhingraFPCDC19}
Bhuwan Dhingra, Manaal Faruqui, Ankur~P. Parikh, Ming{-}Wei Chang, Dipanjan
  Das, and William~W. Cohen. 2019.
\newblock \href {https://doi.org/10.18653/v1/p19-1483} {Handling divergent
  reference texts when evaluating table-to-text generation}.
\newblock In \emph{Proceedings of the 57th Conference of the Association for
  Computational Linguistics, {ACL} 2019, Florence, Italy, July 28- August 2,
  2019, Volume 1: Long Papers}, pages 4884--4895. Association for Computational
  Linguistics.

\bibitem[{DiMarco et~al.(2007)DiMarco, Covvey, Cowan, DiCiccio, Hovy, Lipa,
  Mulholland et~al.}]{dimarco2007development}
Chrysanne DiMarco, HDominic Covvey, D~Cowan, V~DiCiccio, E~Hovy, J~Lipa,
  D~Mulholland, et~al. 2007.
\newblock The development of a natural language generation system for
  personalized e-health information.
\newblock In \emph{Medinfo 2007: Proceedings of the 12th World Congress on
  Health (Medical) Informatics; Building Sustainable Health Systems}, page
  2339. IOS Press.

\bibitem[{ElSahar et~al.(2018)ElSahar, Gravier, and
  Laforest}]{DBLP:conf/naacl/ElSaharGL18}
Hady ElSahar, Christophe Gravier, and Fr{\'{e}}d{\'{e}}rique Laforest. 2018.
\newblock \href {https://aclanthology.info/papers/N18-1020/n18-1020} {Zero-shot
  question generation from knowledge graphs for unseen predicates and entity
  types}.
\newblock In \emph{Proceedings of the 2018 Conference of the North American
  Chapter of the Association for Computational Linguistics: Human Language
  Technologies, {NAACL-HLT} 2018, New Orleans, Louisiana, USA, June 1-6, 2018,
  Volume 1 (Long Papers)}, pages 218--228.

\bibitem[{Gardent et~al.(2017)Gardent, Shimorina, Narayan, and
  Perez{-}Beltrachini}]{DBLP:conf/inlg/GardentSNP17}
Claire Gardent, Anastasia Shimorina, Shashi Narayan, and Laura
  Perez{-}Beltrachini. 2017.
\newblock \href {https://aclanthology.info/papers/W17-3518/w17-3518} {The
  webnlg challenge: Generating text from {RDF} data}.
\newblock In \emph{Proceedings of the 10th International Conference on Natural
  Language Generation, {INLG} 2017, Santiago de Compostela, Spain, September
  4-7, 2017}, pages 124--133.

\bibitem[{Gatt and Krahmer(2018)}]{gatt2018survey}
Albert Gatt and Emiel Krahmer. 2018.
\newblock Survey of the state of the art in natural language generation: Core
  tasks, applications and evaluation.
\newblock \emph{Journal of Artificial Intelligence Research}, 61:65--170.

\bibitem[{Ghazvininejad et~al.(2018)Ghazvininejad, Brockett, Chang, Dolan, Gao,
  Yih, and Galley}]{DBLP:conf/aaai/GhazvininejadBC18}
Marjan Ghazvininejad, Chris Brockett, Ming{-}Wei Chang, Bill Dolan, Jianfeng
  Gao, Wen{-}tau Yih, and Michel Galley. 2018.
\newblock \href
  {https://www.aaai.org/ocs/index.php/AAAI/AAAI18/paper/view/16710} {A
  knowledge-grounded neural conversation model}.
\newblock In \emph{Proceedings of the Thirty-Second {AAAI} Conference on
  Artificial Intelligence, (AAAI-18), the 30th innovative Applications of
  Artificial Intelligence (IAAI-18), and the 8th {AAAI} Symposium on
  Educational Advances in Artificial Intelligence (EAAI-18), New Orleans,
  Louisiana, USA, February 2-7, 2018}, pages 5110--5117.

\bibitem[{Hasan and Farri(2019)}]{hasan2019clinical}
Sadid~A Hasan and Oladimeji Farri. 2019.
\newblock Clinical natural language processing with deep learning.
\newblock In \emph{Data Science for Healthcare}, pages 147--171. Springer.

\bibitem[{He et~al.(2017)He, Balakrishnan, Eric, and
  Liang}]{DBLP:conf/acl/HeBEL17}
He~He, Anusha Balakrishnan, Mihail Eric, and Percy Liang. 2017.
\newblock \href {https://doi.org/10.18653/v1/P17-1162} {Learning symmetric
  collaborative dialogue agents with dynamic knowledge graph embeddings}.
\newblock In \emph{Proceedings of the 55th Annual Meeting of the Association
  for Computational Linguistics, {ACL} 2017, Vancouver, Canada, July 30 -
  August 4, Volume 1: Long Papers}, pages 1766--1776.

\bibitem[{Kingma and Ba(2015)}]{DBLP:journals/corr/KingmaB14}
Diederik~P. Kingma and Jimmy Ba. 2015.
\newblock \href {http://arxiv.org/abs/1412.6980} {Adam: {A} method for
  stochastic optimization}.
\newblock In \emph{3rd International Conference on Learning Representations,
  {ICLR} 2015, San Diego, CA, USA, May 7-9, 2015, Conference Track
  Proceedings}.

\bibitem[{Kiros et~al.(2015)Kiros, Zhu, Salakhutdinov, Zemel, Urtasun,
  Torralba, and Fidler}]{DBLP:conf/nips/KirosZSZUTF15}
Ryan Kiros, Yukun Zhu, Ruslan Salakhutdinov, Richard~S. Zemel, Raquel Urtasun,
  Antonio Torralba, and Sanja Fidler. 2015.
\newblock \href {http://papers.nips.cc/paper/5950-skip-thought-vectors}
  {Skip-thought vectors}.
\newblock In \emph{Advances in Neural Information Processing Systems 28: Annual
  Conference on Neural Information Processing Systems 2015, December 7-12,
  2015, Montreal, Quebec, Canada}, pages 3294--3302.

\bibitem[{Konstas and Lapata(2012)}]{DBLP:conf/naacl/KonstasL12}
Ioannis Konstas and Mirella Lapata. 2012.
\newblock \href {http://www.aclweb.org/anthology/N12-1093} {Unsupervised
  concept-to-text generation with hypergraphs}.
\newblock In \emph{Human Language Technologies: Conference of the North
  American Chapter of the Association of Computational Linguistics,
  Proceedings, June 3-8, 2012, Montr{\'{e}}al, Canada}, pages 752--761.

\bibitem[{Konstas and Lapata(2013)}]{DBLP:journals/jair/KonstasL13}
Ioannis Konstas and Mirella Lapata. 2013.
\newblock \href {https://doi.org/10.1613/jair.4025} {A global model for
  concept-to-text generation}.
\newblock \emph{J. Artif. Intell. Res.}, 48:305--346.

\bibitem[{Le and Mikolov(2014)}]{DBLP:conf/icml/LeM14}
Quoc~V. Le and Tomas Mikolov. 2014.
\newblock \href {http://jmlr.org/proceedings/papers/v32/le14.html} {Distributed
  representations of sentences and documents}.
\newblock In \emph{Proceedings of the 31th International Conference on Machine
  Learning, {ICML} 2014, Beijing, China, 21-26 June 2014}, pages 1188--1196.

\bibitem[{Lebret et~al.(2016)Lebret, Grangier, and
  Auli}]{DBLP:conf/emnlp/LebretGA16}
R{\'{e}}mi Lebret, David Grangier, and Michael Auli. 2016.
\newblock \href {http://aclweb.org/anthology/D/D16/D16-1128.pdf} {Neural text
  generation from structured data with application to the biography domain}.
\newblock In \emph{Proceedings of the 2016 Conference on Empirical Methods in
  Natural Language Processing, {EMNLP} 2016, Austin, Texas, USA, November 1-4,
  2016}, pages 1203--1213.

\bibitem[{Liang et~al.(2009)Liang, Jordan, and Klein}]{DBLP:conf/acl/LiangJK09}
Percy Liang, Michael~I. Jordan, and Dan Klein. 2009.
\newblock \href {http://www.aclweb.org/anthology/P09-1011} {Learning semantic
  correspondences with less supervision}.
\newblock In \emph{{ACL} 2009, Proceedings of the 47th Annual Meeting of the
  Association for Computational Linguistics and the 4th International Joint
  Conference on Natural Language Processing of the AFNLP, 2-7 August 2009,
  Singapore}, pages 91--99.

\bibitem[{Liu et~al.(2018)Liu, Wang, Sha, Chang, and
  Sui}]{DBLP:conf/aaai/LiuWSCS18}
Tianyu Liu, Kexiang Wang, Lei Sha, Baobao Chang, and Zhifang Sui. 2018.
\newblock \href
  {https://www.aaai.org/ocs/index.php/AAAI/AAAI18/paper/view/16599}
  {Table-to-text generation by structure-aware seq2seq learning}.
\newblock In \emph{Proceedings of the Thirty-Second {AAAI} Conference on
  Artificial Intelligence, (AAAI-18), the 30th innovative Applications of
  Artificial Intelligence (IAAI-18), and the 8th {AAAI} Symposium on
  Educational Advances in Artificial Intelligence (EAAI-18), New Orleans,
  Louisiana, USA, February 2-7, 2018}, pages 4881--4888.

\bibitem[{Lu et~al.(2009)Lu, Ng, and Lee}]{DBLP:conf/emnlp/LuNL09}
Wei Lu, Hwee~Tou Ng, and Wee~Sun Lee. 2009.
\newblock \href {http://www.aclweb.org/anthology/D09-1042} {Natural language
  generation with tree conditional random fields}.
\newblock In \emph{Proceedings of the 2009 Conference on Empirical Methods in
  Natural Language Processing, {EMNLP} 2009, 6-7 August 2009, Singapore, {A}
  meeting of SIGDAT, a Special Interest Group of the {ACL}}, pages 400--409.

\bibitem[{Ma et~al.(2019)Ma, Yang, Liu, Li, Zhou, and
  Sun}]{DBLP:conf/acl/MaYLLZS19}
Shuming Ma, Pengcheng Yang, Tianyu Liu, Peng Li, Jie Zhou, and Xu~Sun. 2019.
\newblock \href {https://doi.org/10.18653/v1/p19-1197} {Key fact as pivot: {A}
  two-stage model for low resource table-to-text generation}.
\newblock In \emph{Proceedings of the 57th Conference of the Association for
  Computational Linguistics, {ACL} 2019, Florence, Italy, July 28- August 2,
  2019, Volume 1: Long Papers}, pages 2047--2057. Association for Computational
  Linguistics.

\bibitem[{Mikolov et~al.(2013)Mikolov, Sutskever, Chen, Corrado, and
  Dean}]{DBLP:conf/nips/MikolovSCCD13}
Tomas Mikolov, Ilya Sutskever, Kai Chen, Gregory~S. Corrado, and Jeffrey Dean.
  2013.
\newblock \href
  {http://papers.nips.cc/paper/5021-distributed-representations-of-words-and-phrases-and-their-compositionality}
  {Distributed representations of words and phrases and their
  compositionality}.
\newblock In \emph{Advances in Neural Information Processing Systems 26: 27th
  Annual Conference on Neural Information Processing Systems 2013. Proceedings
  of a meeting held December 5-8, 2013, Lake Tahoe, Nevada, United States.},
  pages 3111--3119.

\bibitem[{Novikova et~al.(2017)Novikova, Dusek, and
  Rieser}]{DBLP:conf/sigdial/NovikovaDR17}
Jekaterina Novikova, Ondrej Dusek, and Verena Rieser. 2017.
\newblock \href {https://aclanthology.info/papers/W17-5525/w17-5525} {The {E2E}
  dataset: New challenges for end-to-end generation}.
\newblock In \emph{Proceedings of the 18th Annual SIGdial Meeting on Discourse
  and Dialogue, Saarbr{\"{u}}cken, Germany, August 15-17, 2017}, pages
  201--206.

\bibitem[{Pennington et~al.(2014)Pennington, Socher, and
  Manning}]{DBLP:conf/emnlp/PenningtonSM14}
Jeffrey Pennington, Richard Socher, and Christopher~D. Manning. 2014.
\newblock \href {http://aclweb.org/anthology/D/D14/D14-1162.pdf} {Glove: Global
  vectors for word representation}.
\newblock In \emph{Proceedings of the 2014 Conference on Empirical Methods in
  Natural Language Processing, {EMNLP} 2014, October 25-29, 2014, Doha, Qatar,
  {A} meeting of SIGDAT, a Special Interest Group of the {ACL}}, pages
  1532--1543.

\bibitem[{Peters et~al.(2018)Peters, Neumann, Iyyer, Gardner, Clark, Lee, and
  Zettlemoyer}]{DBLP:conf/naacl/PetersNIGCLZ18}
Matthew~E. Peters, Mark Neumann, Mohit Iyyer, Matt Gardner, Christopher Clark,
  Kenton Lee, and Luke Zettlemoyer. 2018.
\newblock \href {https://aclanthology.info/papers/N18-1202/n18-1202} {Deep
  contextualized word representations}.
\newblock In \emph{Proceedings of the 2018 Conference of the North American
  Chapter of the Association for Computational Linguistics: Human Language
  Technologies, {NAACL-HLT} 2018, New Orleans, Louisiana, USA, June 1-6, 2018,
  Volume 1 (Long Papers)}, pages 2227--2237.

\bibitem[{Puduppully et~al.(2018)Puduppully, Dong, and
  Lapata}]{DBLP:journals/corr/abs-1809-00582}
Ratish Puduppully, Li~Dong, and Mirella Lapata. 2018.
\newblock \href {http://arxiv.org/abs/1809.00582} {Data-to-text generation with
  content selection and planning}.
\newblock \emph{CoRR}, abs/1809.00582.

\bibitem[{Radford et~al.(2019)Radford, Wu, Child, Luan, Amodei, and
  Sutskever}]{radford2019language}
Alec Radford, Jeff Wu, Rewon Child, David Luan, Dario Amodei, and Ilya
  Sutskever. 2019.
\newblock Language models are unsupervised multitask learners.

\bibitem[{Reiter and Dale(1997)}]{DBLP:journals/nle/ReiterD97}
Ehud Reiter and Robert Dale. 1997.
\newblock \href {https://doi.org/10.1017/S1351324997001502} {Building applied
  natural language generation systems}.
\newblock \emph{Natural Language Engineering}, 3(1):57--87.

\bibitem[{Saha et~al.(2018)Saha, Pahuja, Khapra, Sankaranarayanan, and
  Chandar}]{DBLP:conf/aaai/SahaPKSC18}
Amrita Saha, Vardaan Pahuja, Mitesh~M. Khapra, Karthik Sankaranarayanan, and
  Sarath Chandar. 2018.
\newblock \href
  {https://www.aaai.org/ocs/index.php/AAAI/AAAI18/paper/view/17181} {Complex
  sequential question answering: Towards learning to converse over linked
  question answer pairs with a knowledge graph}.
\newblock In \emph{Proceedings of the Thirty-Second {AAAI} Conference on
  Artificial Intelligence, (AAAI-18), the 30th innovative Applications of
  Artificial Intelligence (IAAI-18), and the 8th {AAAI} Symposium on
  Educational Advances in Artificial Intelligence (EAAI-18), New Orleans,
  Louisiana, USA, February 2-7, 2018}, pages 705--713.

\bibitem[{See et~al.(2017)See, Liu, and Manning}]{DBLP:conf/acl/SeeLM17}
Abigail See, Peter~J. Liu, and Christopher~D. Manning. 2017.
\newblock \href {https://doi.org/10.18653/v1/P17-1099} {Get to the point:
  Summarization with pointer-generator networks}.
\newblock In \emph{Proceedings of the 55th Annual Meeting of the Association
  for Computational Linguistics, {ACL} 2017, Vancouver, Canada, July 30 -
  August 4, Volume 1: Long Papers}, pages 1073--1083.

\bibitem[{Sennrich et~al.(2016)Sennrich, Haddow, and
  Birch}]{DBLP:conf/acl/SennrichHB16a}
Rico Sennrich, Barry Haddow, and Alexandra Birch. 2016.
\newblock \href {http://aclweb.org/anthology/P/P16/P16-1162.pdf} {Neural
  machine translation of rare words with subword units}.
\newblock In \emph{Proceedings of the 54th Annual Meeting of the Association
  for Computational Linguistics, {ACL} 2016, August 7-12, 2016, Berlin,
  Germany, Volume 1: Long Papers}.

\bibitem[{Su et~al.(2016)Su, Sun, Sadler, Srivatsa, Gur, Yan, and
  Yan}]{DBLP:conf/emnlp/SuSSSGYY16}
Yu~Su, Huan Sun, Brian~M. Sadler, Mudhakar Srivatsa, Izzeddin Gur, Zenghui Yan,
  and Xifeng Yan. 2016.
\newblock \href {http://aclweb.org/anthology/D/D16/D16-1054.pdf} {On generating
  characteristic-rich question sets for {QA} evaluation}.
\newblock In \emph{Proceedings of the 2016 Conference on Empirical Methods in
  Natural Language Processing, {EMNLP} 2016, Austin, Texas, USA, November 1-4,
  2016}, pages 562--572.

\bibitem[{Walker et~al.(2001)Walker, Rambow, and
  Rogati}]{DBLP:conf/naacl/WalkerRR01}
Marilyn~A. Walker, Owen Rambow, and Monica Rogati. 2001.
\newblock \href {http://aclweb.org/anthology/N/N01/N01-1003.pdf} {Spot: {A}
  trainable sentence planner}.
\newblock In \emph{Language Technologies 2001: The Second Meeting of the North
  American Chapter of the Association for Computational Linguistics, {NAACL}
  2001, Pittsburgh, PA, USA, June 2-7, 2001}.

\bibitem[{Wiseman et~al.(2017)Wiseman, Shieber, and
  Rush}]{DBLP:conf/emnlp/WisemanSR17}
Sam Wiseman, Stuart~M. Shieber, and Alexander~M. Rush. 2017.
\newblock \href {https://aclanthology.info/papers/D17-1239/d17-1239}
  {Challenges in data-to-document generation}.
\newblock In \emph{Proceedings of the 2017 Conference on Empirical Methods in
  Natural Language Processing, {EMNLP} 2017, Copenhagen, Denmark, September
  9-11, 2017}, pages 2253--2263.

\bibitem[{Wiseman et~al.(2018)Wiseman, Shieber, and
  Rush}]{DBLP:conf/emnlp/WisemanSR18}
Sam Wiseman, Stuart~M. Shieber, and Alexander~M. Rush. 2018.
\newblock \href {https://aclanthology.info/papers/D18-1356/d18-1356} {Learning
  neural templates for text generation}.
\newblock In \emph{Proceedings of the 2018 Conference on Empirical Methods in
  Natural Language Processing, Brussels, Belgium, October 31 - November 4,
  2018}, pages 3174--3187.

\bibitem[{Yin et~al.(2016)Yin, Jiang, Lu, Shang, Li, and
  Li}]{DBLP:conf/ijcai/YinJLSLL16}
Jun Yin, Xin Jiang, Zhengdong Lu, Lifeng Shang, Hang Li, and Xiaoming Li. 2016.
\newblock \href {http://www.ijcai.org/Abstract/16/422} {Neural generative
  question answering}.
\newblock In \emph{Proceedings of the Twenty-Fifth International Joint
  Conference on Artificial Intelligence, {IJCAI} 2016, New York, NY, USA, 9-15
  July 2016}, pages 2972--2978.

\end{thebibliography}
\bibliographystyle{acl_natbib}

\appendix

\section*{Appendix A. Implementation Details}
We use the Adam optimizer~\cite{DBLP:journals/corr/KingmaB14} with learning rate set to 0.0003. The mini-batch size is set to 40 and the weight $\lambda$ of the copy loss term to 0.7. The dimension of the position embedding is set to 5. For attribute name with multiple words, we average their word embeddings as the attribute name embedding. Refer to our released code and data at \url{https://github.com/czyssrs/Few-Shot-NLG} for more details.

\section*{Appendix B. ROUGE-4 Results}
Following previous work~\cite{DBLP:conf/aaai/LiuWSCS18}, we conduct automatic evaluations using BLEU-4 and ROUGE-4 (F-measure)\footnote{We use standard scripts NIST mteval-v13a.pl (for BLEU), and
rouge-1.5.5 (for ROUGE)}. Table~\ref{table:res_rouge_humans}, ~\ref{table:res_rouge_books} and ~\ref{table:res_rouge_songs} show the ROUGE-4 results for three domains \textit{Humans}, \textit{Books} and \textit{Songs}, respectively.  
\begin{table}[htbp]
\begin{center}
\resizebox{0.48\textwidth}{!}{%
\begin{tabular}{lccccc}
\toprule
Domain &
\multicolumn{4}{c}{Humans}\\
\cline{2-6}
\# of training instances & - & 50 & 100 & 200 & 500\\
\midrule
Template & 5.1 & - & - & - & - \\
\midrule
Base-original & - & 0.1 & 0.4 & 0.5 & 0.6\\
Base & - & 0.1 & 0.4 & 0.8 & 1.5\\
\midrule
Base+switch & - & 4.9 & 6.3 & 9.8 & 12.5\\
Base+switch+LM-scratch & - & 1.0 & 2.8 & 4.7 & 7.1\\
Base+switch+LM (Ours) & - & \textbf{14.1} & \textbf{16.2} & \textbf{22.1} & \textbf{28.3}\\
\bottomrule
\end{tabular}
}
\end{center}
\caption{ROUGE-4 results on \textit{Humans} domain}
\label{table:res_rouge_humans}
\end{table}

\begin{table}[htbp]
\begin{center}
\resizebox{0.48\textwidth}{!}{%
\begin{tabular}{lccccc}
\toprule
Domain &
\multicolumn{4}{c}{Books}\\
\cline{2-6}
\# of training instances & - & 50 & 100 & 200 & 500\\
\midrule
Template & 15.0 & - & - & - & - \\
\midrule
Base-original & - & 1.1 & 1.6 & 2.1 & 1.5\\
Base & - & 1.7 & 1.5 & 2.1 & 2.4\\
\midrule
Base+switch & - & 12.8 & 15.0 & 18.1 & 20.7\\
Base+switch+LM-scratch & - & 2.4 & 4.2 & 6.5 & 10.7\\
Base+switch+LM (Ours) & - & \textbf{22.5} & \textbf{23.1} & \textbf{25.0} & \textbf{27.6}  \\
\bottomrule
\end{tabular}
}
\end{center}
\caption{ROUGE-4 results on \textit{Books} domain}
\label{table:res_rouge_books}
\end{table}

\begin{table}[htbp]
\begin{center}
\resizebox{0.48\textwidth}{!}{%
\begin{tabular}{lccccc}
\toprule
Domain &
\multicolumn{4}{c}{Songs}\\
\cline{2-6}
\# of training instances & - & 50 & 100 & 200 & 500\\
\midrule
Template & 24.5 & - & - & - & - \\
\midrule
Base-original & - & 3.4 & 4.2 & 4.7 & 4.8\\
Base & - & 4.1 & 5.1 & 4.7 & 5.8\\
\midrule
Base+switch & - & 20.2 & 21.7 & 23.2 & 24.8\\
Base+switch+LM-scratch  & - & 5.4 & 8.0 & 12.0 & 15.0\\
Base+switch+LM (Ours) & - & \textbf{26.2} & \textbf{28.6} & \textbf{30.1} & \textbf{32.6}\\
\bottomrule
\end{tabular}
}
\end{center}
\caption{ROUGE-4 results on \textit{Songs} domain}
\label{table:res_rouge_songs}
\end{table}

\section*{Appendix C. Human Evaluation Details}
We conduct human evaluation studies using Amazon Mechanical Turk, based on two aspects: \textit{Factual correctness} and \textit{Language naturalness}. For both studies, we evaluate the results trained with 200 training instances of \textit{Humans} domain. We randomly sample 500 instances from the test set, together with the texts generated with different methods. Each evaluation unit is assigned to 3 workers to eliminate human variance. 

The first study attempts to evaluate how well a generated text can correctly convey information in the table. Each worker is present with \textit{both the input table and a generated text}, and asked to count how many facts in the generated text are supported by the table, and how many are contradicting with or missing from the table, similar as in~\cite{DBLP:conf/emnlp/WisemanSR17}. The we calculate the average number of supporting and contradicting facts for the texts generated by each method.

The second study aims to evaluate whether the generated text is grammatically correct and fluent in terms of language, regardless of factual correctness. Each worker is present with a pair of texts generated from the same input table, by two different methods, then asked to select the better one only according to language naturalness, or ``Tied" if the two texts are of equal quality. \textit{The input table is not shown to the workers}. Each time a generated text is chosen as the better one, we assign score of 1.0. If two texts are tied, we assign 0.5 for each. We then calculate the average score for the texts generated by each method, indicating its superiority in pairwise comparisons with all other methods. 

The significance test is conducted respectively on all three measures: number of supporting facts and number of contradicting facts for the first study; the assigned score for the second study. We use the Tukey HSD post-hoc analysis of an ANOVA with the worker's response as the dependent variable, the method and worker id as independent variables.

\end{document}